# Ear Recognition

Nikolaos Athanasios Anagnostopoulos

**Introduction to Biometrics**

University of Twente

EIT ICT Labs Master School

**2013-14**

UNIVERSITEIT TWENTE.

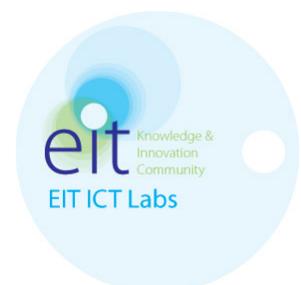

# Abstract

Ear recognition can be described as a revived scientific field. Ear biometrics were long believed to not be accurate enough and held a secondary place in scientific research, being seen as only complementary to other types of biometrics, due to difficulties in measuring correctly the ear characteristics and the potential occlusion of the ear by hair, clothes and ear jewellery. However, recent research has reinstated them as a vivid research field, after having addressed these problems and proven that ear biometrics can provide really accurate identification and verification results.

Several 2D and 3D imaging techniques, as well as acoustical techniques using sound emission and reflection, have been developed and studied for ear recognition, while there have also been significant advances towards a fully automated recognition of the ear. Furthermore, ear biometrics have been proven to be mostly non-invasive, adequately permanent and accurate, and hard to spoof and counterfeit. Moreover, different ear recognition techniques have proven to be as effective as face recognition ones, thus providing the opportunity for ear recognition to be used in identification and verification applications.

Finally, even though some issues still remain open and require further research, the scientific field of ear biometrics has proven to be not only viable, but really thriving.

# 1 Introduction

Ear recognition traces its roots to French criminologist A. Bertillon,[1][2] who in an 1890 seminar supported that the ear is an important factor for identification and recognition.[3] In the United States of America, an ear classification system based on manual measurements was developed by A. V. Iannarelli, and ear recognition has played some role in the forensic science for the past 40 years.[1][2]

Iannarelli proposed a biometric recognition system based on twelve measurements of ear characteristics. Photographs of the right ear were to be taken and then registered, after being aligned and scaled during development.[4] This allowed the photographs to be normalised in size and orientation and therefore be easily comparable to each other. Measurements of the distance between the different ear characteristics made in units of 3mm were stored along with information on sex and race,[4] thus creating relatively unique sets of identification characteristics. However, due to the importance of identifying the exact points of measurements in great detail, this method has been generally considered unsuitable for machine vision.

Over the years, a high number of different techniques has been introduced for ear recognition, identification and verification. However, significant difficulties against the correct and adequate acquisition of ear biometrics rising from the difficulty to measure the ear's anatomy details and the easy and quite common concealment of the area of the ears have persisted. This has led ear biometrics to have a minor and only complementary role in identification and verification, including any forensic science applications.

## 2 The difficulties of ear recognition

Apart from the considerable difficulties of correctly measuring the details of the anatomy of the ear, there are considerable impediments in the process of acquiring suitable pictures, schematics or photographs of the ear. Besides the obvious hindrances that concealment of the ear by clothes, hair, ear ornaments and jewellery may pose, the issue of correct registration of the ear biometrics acquired through analog or digital images remains valid, as the availability or not of images taken at different angles may also conceal or disclose important characteristics of the ear's anatomy.

However the employment of ear ornaments and jewellery may also provide unique ear identifiers that can prove critical in the field of forensic science. The temporary and volatile nature of such ornaments may not allow their utilisation as identifiers for long time periods, but in many cases this could be compensated by exploitation of the related ear markings, left from the usage of such jewellery and ornaments, especially in such cases as ear lobe stretchers.

Nevertheless, the problem of appropriate and sufficient acquisition of the ear's image and characteristics is further exacerbated by such elements as the presence of hair or clothes covering all or part of the ear in a great number of images acquired during unexceptional circumstances.

These difficulties have led ear recognition to be playing a secondary role in identification systems and techniques, being complementary to other biometrics, which are being used more commonly for positive identification and verification. Recently, however, there is a renewed interest in the field of ear recognition, which may be potentially attributed to the advancement of capturing techniques and capabilities, which could compensate for the shortcomings of acquired images. Furthermore, significant process has been made regarding the editing, manipulation and refinement of captured images, which could help reconstruct any obscure or concealed parts or inadequately captured characteristics of the ear with significant success.

A growing number of different approaches is being applied in the field of ear recognition. These approaches include the application of both optical and acoustical means for ear recognition, while they may also involve 2D as well as 3D identification techniques.[5] The existence and employment of these different approaches has not only been demonstrated and reported theoretically, but has also led into an actual number of related patents being granted.[5] This fact can again be used to signify the revived interest around ear biometrics and their applications.

In turn, this has led into increased recent research over new means, ways and techniques that could potentially improve, or even revolutionise, the field of ear recognition and identification. However, it should again be noted that such research seems to suggest that ear biometrics may provide better results when combined with other techniques of identification, such as face and iris recognition, significantly enhancing and improving them.[6][7]

## 3 Recent developments in ear biometrics

As mentioned before, ear biometrics involve a relatively high amount of issues that have to be dealt with before the ear can be successfully employed for identification and verification. These issues include potential concealment of the ear, pictures and schematics produced at angles which conceal important ear features and characteristics, the effects of age,[8] as well as ear markings caused by ear jewellery and ornaments. On top of these issues, if optical 2D ear recognition was to be employed in a widespread fashion, such countermeasures as deliberate partial or full concealment of the ear would prove to be highly effective. For this purpose, ear biometrics have at best been employed in only a complementary fashion to other means of recognition, identification and verification, especially in forensic science applications.

However research on acoustical ear recognition may help overcome the obstacles described. Acoustical ear recognition uses sounds reflected by or on the ear during hearing and could potentially be used even when the ear is partially or fully concealed. As sound can penetrate quite well thin objects, such as hair, clothes or other items that usually cover ears, it can be used to either form a picture of the ear and its surface characteristics or to measure the response of the ear to it.[2][9]

Even though sound scattering on a concealed ear may not give an adequately clear image of the ear, due to the ear's relatively complex structure, the variety of frequencies that can be employed and the different angles that can be used may make it possible to reconstruct a good enough image of the ear, based on capturing the sounds that are reflected by it. It is important to note that even if the a full ear image cannot be reconstructed to high detail, it may still be able to measure the position of its basic features with enough detail to allow for them to be used as a biometric. It is also interesting to mention here that the same technique could be employed for face recognition as well, when features such as the nose and its tip, the chin and the cheeks are considered.

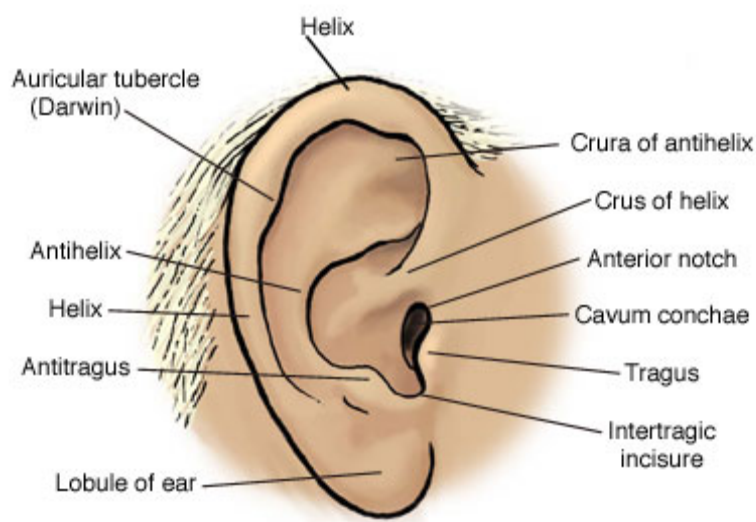

**Fig. 1**: *2D sketch of the external ear anatomy.*[10]

Furthermore, the way sound is reflected by the ear is quite unique due to the unique construction and combination of the outer ear structure, the ear canal and the ear drum. Therefore, this can also be used as an identifier with quite good results.[11] In addition to this, general otoacoustic emissions, sounds that the ear emits either due to external stimulation or on its own during the process of hearing, are also unique and recent studies have shown their potential for being used as a biometric modality.[12][13]

Moreover, all ear recognition, identification and verification techniques can take advantage of the presence of two ears on most humans, which can provide room for techniques that may compare features of the left and the right ear against each other or may combine the biometrics of each ear in addition to each other, so that these techniques may become more reliable. It should be noted here that unlike, for example, eyes, the left and the right ear of a single person are not identical, or even so highly similar to each other. Each ear has a significant degree of uniqueness,[11] with features such as the helix differing in a significant degree between the left and the right ear of the same person.[14]

Finally, 3D imaging techniques seem to provide another way to provide highly reliable ear biometrics that can be used for successful recognition and identification.[15] Recent studies have shown a rank-1 recognition rate of around 95% for such 3D imaging techniques for ear biometrics.[16][17] Even though 3D techniques do not provide significantly more additional ear biometrics to 2D imaging, they can, nevertheless, raise the accuracy of the characteristics being captured and thus lead into better registration.[14] However, these techniques tend to be highly dependent on the illumination and shading of the captured images and require preprocessing to remove artifacts introduced by the 3D sensor. [17][18]

It has to be mentioned, however, that the best results are always achieved when ear biometrics are combined with other relevant biometrics, such as iris, face or fingerprints. A growing number of studies has shown that combining ear biometrics with another type of biometrics will lead to better results when compared with any of the two types of biometrics alone.[6][7][18][19][20]

# 4 The significance of ear biometrics

Recent research has shown that, contrary to what has long been believed, ear biometrics do not perform worse than other biometrics, such as face, towards positive identification and verification.[19] Although an initial study comparing ear recognition and face recognition using Principal Component Analysis (PCA) showed that face recognition at the time was performing better,[20] more recent techniques have allowed ear recognition to reach very high recognition rates, similar to those attained by face recognition.

It has also been empirically proven through research that ears are relatively unique.[9][21] Furthermore, even though age affects ears in a significant way,[8] their characteristics are relatively more permanent than, for example, facial characteristics and at the same level of permanency as fingertips.[21] These qualities contribute into ears being relatively suitable biometrics for identification and verification.

These qualities combined with the relatively low cost of the methods and techniques used for the acquisition and handling of ear biometrics,[21][22] make ear biometrics particularly interesting for further research. Furthermore, it must be noted that ear biometrics involve a relatively non-intrusive acquisition, as long as the area of the ear is not fully covered. However, the same holds true for most biometrics; as the area associated with their acquisition should be not covered.

Moreover, ear biometrics are relatively difficult to counterfeit because of the ear's deep 3D structure and their robust resistance to change with age, and therefore can be used for identification and verification purposes when a high degree of protection is required.[2] However, earmarks left behind are not of the same quality as fingertips and cannot be used conclusively for identification or verification. Nevertheless, ear biometrics could also be used in recognition and verification with less effort required from the person that wants to be identified or verified, as their use doesn't involve them being remembered or possessed in such ways as passwords and cards.

Even though ear biometrics are not currently used as a primary means of identification and/or verification, there is concrete evidence that they are as suitable and reliable as face biometrics and therefore there is no reason why they should not be used more widely in the future. Comparisons of recent 2D techniques and methods have consistently demonstrated an accuracy of over 90%, while 3D techniques achieve an accuracy of over 95%.[9] Furthermore, the combination of ear biometrics with other types of biometrics has a consistent accuracy of over 95%.[9]

Moreover, different techniques have employed light and sound of a specific frequency to deal with the problem of concealed or obscure ears, with significant success. Such a technique uses infra-red light to detect ear characteristics in conditions of relative darkness.[23] Ear biometrics obtained in such a way can be used to achieve identification or verification as the examined characteristics do not vary by colour, unlike, for example, eyes.

Another interesting fact is that, as ear biometrics has not been widely used in recognition, identification and verification systems, they are still relatively unknown to potential culprits and there has not yet been any significant attempt to spoof them. Therefore, and taking into account the relative difficulty of counterfeiting them, they could be used in forensic science and in the judicial system, in a better way than other techniques already used such as fingertips. It should however be noted that the area of the ear's helix is relatively elastic and could be deliberately deformed to deceive an ear recognition system. However, in a similar fashion to the elasticity of human fingerprints or facial characteristics, once those deliberate deformities have been identified, they can be digitally addressed, and the identification system can be trained to recognise and correct them. Additionally, other techniques exist which can revert the results of such deliberate elastic deformation of the ear by utilising different sets of ear biometrics.[24]

Finally, even though there has some effort to classify ear identification methods into different categories, usually according to the principal characteristic or technique used, for all 2D, 3D and acoustical techniques, such classifications are not standardised,

widespread or conclusive yet.[5][9]

# 5 Current methods of ear recognition

It should be noted that automated ear recognition techniques do exist at present, based on machine vision.[4] These techniques, however, face the same challenges as other computer vision techniques, sometimes failing to correctly identify the whole ear region and so on. It is evident, though, that ear registration or recognition processes can be further sped up and automated, if efficient and dependable algorithms can be used for the automatic detection of ear landmarks and contours.[1]

Different window shapes for the capturing of the ear region have been proposed, while also capturing whole blocks of the face containing the ear area has been suggested.[2][25] Window shapes that have been proposed include both circular and square approaches. In relation to this, it must be noted that Iannareli classified ears into the following distinct shape types: round, rectangular, triangular, oval, or undetermined,[26] which could provide helpful insight about the shape of the window to be used.

Other types of preliminary classification include levelling according to concealment of the ear region and using colour or other characteristics as classifiers, such as whether the ear lobule is detached or attached.[27] Using such characteristics during registration can improve the accuracy and speed of a system during identification and verification,[27] but of course may also increase the time required for registration.

Furthermore, other techniques combine the characteristics of the ear with other face features, using the position of the ear, or its more specific features, relative to each side's eye brow, cheek or eye, sometimes in combination with the position of the nose (tip), as a single combinatorial biometric. Also, the position of the ear in relation to the nose tip alone could be used as a biometric on its own. Such systems combining different face characteristics and the ear may be more accurate than already existing ear biometrics, depending on the correct and accurate registration of all features involved.

Moreover, ear registration techniques may adapt the use of holistic and/or local descriptors,[1] depending on issues of pattern recognition, such as the local texture of the whole or parts of the ear.[5] The triangulation of the area of the ear has also been tested for the recognition and registration for the ear and its characteristics, with quite good results, providing high quality 3D ear images in real time.[28]

An initial technique employed for 2D ear recognition involved using Principal Component Analysis (PCA) in the same way as face recognition, but replacing eigen-faces with eigen-ears. However, a large amount of other methods and techniques was relatively soon introduced in the field of ear biometrics in an effort for improved results and accuracy.[1][9][22] In 3D ear biometrics, techniques such as ICP (Iterative Closest Point) and wavelet transformation have provided good results and quite accurate identification and verification.[1][9][18][29]

Finally, the ear can also be recognised and registered through sound, but again there must be sufficient registration, performed in an adequate way, which would avoid external noise being registered. Furthermore, external noise during the process of recognition must also be removed in an efficient way.

# 6 Novel approaches in 2D ear recognition and their performance

Various 2D ear recognition techniques have recently been used to provide results with an accuracy of over 75%, which usually tends to be around 90%.[9] Furthermore, a classification of such techniques based on their approach has been proposed, but is not yet standardised or even widely accepted.[5] Based on it, 2D ear recognition techniques are classified into four categories, utilising a structural, a subspace learning, a model-based or a spectral approach.[5]

However, different approaches in classification of ear recognition techniques exist, which group them according to the actual method used for recognition.[1][9] Based on these classifications, we can again note the high accuracy ratio illustrated, as well as focus on some examples of demonstrated techniques and the exact results they have provided. In one of these classifications,[9] recent 2D ear recognition methods are placed in the following categories: intensity-based, Force Field (FF), Fourier descriptor, wavelet transformation, Gabor filters and SIFT (Scale-Invariant Feature Transform) techniques.

Intensity-based techniques, which utilise a subspace learning approach, include PCA, ICA (Independent Component Analysis), FSLDA (Full-Space Linear Discriminant Analysis), IDLLE (Improved Locally Linear Embedding), NKDA (Null Kernel Discriminant Analysis) and Sparce Representation.[5][9] Specifically, a PCA technique described by Chang, Bowyer, Sarkar and Victor in 2003,[19] using a training set of 197 subjects, with 111 subjects for lighting variation experiments and 101 subjects also used for pose variation experimental purposes, had a recognition performance of more than 70% for day variation, more than 60% for lighting variation and more than 20% for pose variation. The researchers also noted that they did not find any significant difference in the recognition performance between the face and the ear, noting that, in one experiment, this was 70.5% percent for face versus 71.6% for the ear.[19] They also found that multimodal recognition using both the ear and face resulted in statistically significant improvement over either individual biometric.[19]

An ICA experiment described by Zhang, Mu, Qu, Liu and Zhang in 2005,[30] included 17 subjects who provided 6 ear images each which were cropped and rotated for uniformity, as well as 60 subjects who provided 3 ear images, which were again cropped, but not rotated or brightened. It demonstrated a maximum classification accuracy of 88.23% by using PCA on the first set and of 85% by using PCA on the second set, while it also showed a maximum accuracy of 94.11% by using ICA on the first set and of 88.33% by using ICA on the second set.[30]

An FSLDA technique was applied by Yuan and Mu in 2007 for ear recognition, after an automatic ear normalization method based on improved Active Shape Model (ASM).[31] 79 subjects were used providing ear images taken from 9-10 different angles. Depending on the angle, the recognition rates are usually over 90%, dropping to 80% or much less as the angle drops at 20 degrees or lower.[31]

An improved LLE method is introduced by Xie and Mu in 2008 and tested against 16 datasets of 79 ear images each.[32] Depending on the angle of the images provided, recognition rates are proven to rise above 80% for angles below 20 degrees and well above 90% for angles of more degrees.[32]

A new multi-view ear recognition technique based on B-Spline pose manifold construction in discriminative projection space which is formed by null kernel discriminant analysis (NKDA) feature extraction was presented by Zhang and Liu in 2008.[33] 60 subjects were used and different techniques were compared to the one proposed, resulting in rank-1 recognition rates between 75% and 80% for PCA and well over 97% for the novel technique.[33]

Another technique based on Sparce Representation was proposed in 2008 by Naseem, Togneri and Bennamoun.[34] 32 subjects provided 6 ear images each, taken under varying lighting conditions and with the head rotations of -90 and -75 degrees.[34] Two different algorithms implementing this technique resulted in recognition rates of 91.67% and 96.88%.[34]

SIFT is another ear recognition technique based on a subspace learning approach. Kisku, Mehrotra, Gupta and Sing published a technique based on SIFT in 2009, which based on whether Euclidean distances or nearest neighbour metrics are used can provide an accuracy of 91.09% and 93.01%, respectively.[35] 400 subjects are used in this technique, providing one samples each for registration and one for testing. Additionally, if colour segmentation is also used, then the accuracy rates rise to 94.31% and 96.93%, respectively.[35]

Techniques based on Gabor filters, which implement a spectral approach, have also provided quite efficient results regarding ear recognition. A technique proposed by Nanni and Lumini in 2009 provides a rank-1 recognition rate of 84.6%, based on experiments on 464 ear images obtained from 114 users, on different days, with different conditions of pose and lighting, consisting of sets of 3–9 samples from each user.[36] This publication also clearly discusses the various potential obstacles that need to be overcome for successful ear recognition, such as ear occlusion.[36]

A technique that was developed by Watabe, Sai, Sakai and Nakamura in 2008 and extends the idea of elastic graph matching by employing Gabor jets has a rank-1 recognition rate of about 98% as demonstrated in experiments on a database of 362 samples obtained by 181 subjects.[24] Each subject provided one sample for registration and another for testing.[24]

Another technique based on a spectral approach utilises Fourier descriptors and was described by Abate, Nappi, Riccio and Ricciardi in 2006.[37] The researchers have made use of two distinct datasets. The first dataset contains 2 sessions with 210 ear images from 70 peoples, 3 for each subject, with different rotation angles. The second dataset consists of 2 sessions with 72 ear images from 36 peoples, 2 photos per subject looking up with a free rotation angle. Ear images from the first dataset have been manually normalized, while an automatic ear detector was applied to the second dataset in order to extract the ears.[37] While for the first dataset, this method provides an 88% to 96% rank-1 recognition rate, depending on the rotation angle, for the second dataset, it only provides a rank-1 rate of about 62%.[37] However, in all cases, these recognition rates are better than the ones achieved by using PCA on the same datasets.[37]

Techniques based on a model-based approach, such as the one described by Nanni and Lumini in 2007,[38] can also provide a rank-1 recognition rate of 70% to 80%, depending on the method used. The dataset again consisted of 464 ear images obtained from 114 users, on different days, with different conditions of pose and lighting, in sets of 3–9 samples from each user.[38]

Another class of techniques, making use again of a spectral approach, are based on wavelets. Techniques implementing wavelet transformation have been described to provide significantly high recognition rates. For example, a technique published by Sana and Gupta in 2007 has provided accuracy of 97.8% and 98.2%, for a dataset of 3 samples per individual provided by 600 individuals and a dataset of 3 samples per individual provided by 350 individuals, respectfully.[39]

Another example is a technique introduced by Zhao and Mu in 2009, which describes a new approach under which low frequency sub-images are obtained by utilizing a two-dimensional wavelet transform and then the features are extracted by applying Orthogonal Centroid Algorithm to them.[40] This technique is applied to the USTB(77) ear database, which consists of images from 77 different people, using 4 images from each person, for a total of 308 images, and the USTB(79) ear database, which consists of images from 79 different people, using 11 images from each person, for a total of 869 images.[40] For each database this technique provides better recognition rates than PCA with LDA (Linear Discriminant Analysis), however recognition rates for the new technique may range from 64.94% to 100% for USTB(77) and between 83.54% and 100% for USTB(79).[40]

Furthermore, Nosrati, Faez and Faradji in 2007 proposed combining a 2D wavelet approach with PCA in order to get recognition ranks of 90.5% for a USTB database of 308 images (the USTB(77) described above) and between 95.05% and 97.05% for the Carreira-Perpinan database, which consists of 102 grayscale images (6 images for each of 17 subjects) in PGM format.[41] The difference in the recognition rates of the same database has to do with the number of images per person that are used for registration and training purposes; in the case of a recognition rate of 95.05%, 3 images were used, while in the case of a rate of 97.05%, 4 images were used for these purposes.[41]

In 2008, Wang, Mu and Zeng published another novel method based on Haar wavelet transform and uniform local binary patterns (ULBPs).[42] This technique was tested on a dataset of 79 subjects, each providing images at five different angles.[42] Two images per angle were provided and the angles were set at 0, 5, 20, 35 and 45 degrees.[42] While the recognition rate for angles up to 20 degrees was above 92%, as the angle grew wider, the recognition rate started deteriorating rapidly.[42] Similar results have also been described before regarding the relation of other techniques (e.g. FSLDA, LLE) to the angle at which the images of the ear have been taken.

Finally, structural approaches seem to be mostly used nowadays for ear detection purposes, rather than ear recognition.[9][43] However, as it has already been mentioned, these were the initial approaches on which ear recognition was based upon and therefore, form not only the basis of ear detection, but also of ear recognition. In the future, it may be possible to either draw new ideas for more adequate ear recognition or significantly improve ear detection by focusing on the structure of the ear.

# 7 Conclusions

Ear biometrics could easily be described as a re-discovered field of biometrics. While it was believed that ear biometrics may not be accurate enough, it has now been clearly demonstrated that this is not the case. Ear biometrics were initially hindered by such inherent problems as the partial or full concealment of the ear by hair, clothes, or ear ornaments and jewellery, and the difficulty of correctly measuring the ear's characteristics. These problems had caused ear biometrics to play a secondary role, being used only complementary to other biometrics at best.

However, recent research has managed to address such problems in a satisfactory manner, while also proving that ear recognition can lead to accurate identification and verification. This, in turn, has led to a revived interest in further research of the ear biometrics, which could improve, or even revolutionise, the field of ear recognition and identification.

Several different techniques have been developed for ear recognition, employing 2D and 3D images, models, pictures and schematics, as well as the acoustics of the ear through sound emission and reflection. These techniques can already compensate in a significant degree for ear occlusion and other relevant problems, and may in the future lead into reasonably better results and accuracy in ear identification and verification.

Regarding 2D ear recognition in particular, various different techniques have been employed providing, in general, quite high recognition rates and accuracy of more than 75% and which tend to be above 90% in the most optimum cases. These methods and techniques could generally be classified according to the approach they use to achieve ear recognition. Based on this, four different categories of 2D ear recognition have been identified, classifying 2D ear recognition techniques into ones utilising structural characteristics, spectral approaches, using subspace learning or employing models in order to correctly identify an ear.

Among the most commonly used are subspace learning approaches, in techniques based upon PCA, ICA, LDA, LLE, NKDA, Sparce Representation and SIFT, and spectral approaches, utilising Gabor filters, Gabor jets, Fourier descriptors and wavelets. Structural characteristics seem to be used more for ear detection than for actual identification. While both the techniques that are based on subspace learning and the techniques that take a spectral approach towards ear recognition can provide recognition rates well above 90%, reaching even above 95% in their most optimum settings.

However, the recognition rates of techniques based on either approach deteriorate really fast as the angle at which the 2D ear image was taken increases. For angles up to 20 degrees, recognition rates are quite high, but for angles larger than 20 degrees, they start to rapidly fall below an adequate level. As this could be caused by certain needed ear characteristics being no longer visible and thus measurable, perhaps some techniques of reconstructing the ear or collecting and registering more samples could greatly compensate for this and help improve ear recognition even in this case.

As already mentioned before, even though initial studies proved face recognition to be more effective than ear recognition, it has now been adequately shown that ear biometrics can be as accurate and have as high recognition rates as face biometrics. When comparing the FRR (False Rejection Rate) at the same FAR (False Acceptance Rate) between recent novel ear recognition methods[24] and the latest large-scale results regarding face recognition (namely from FRVT (Face Recognition Vector Test) 2006),[44] these are comparable and of the same magnitude, if not almost exactly similar.

Specifically, both ear recognition and face recognition techniques can give around 1% FRR at 0.1% FAR.[24][39][44] The same holds true regarding recognition rates.[19][30][45] It is not clear whether face recognition still holds a small advantage over ear biometrics, but since the research effort, time and resources assigned to face recognition over the years clearly exceeds that placed on ear recognition, this a small advantage of face recognition results against ear recognition results can be attributed to this fact. However, the ear exhibits less variability with expressions and orientation, while having a more uniform distribution of color intensity and spatial resolution than faces.[32][35] Finally, 3D ear recognition techniques can significantly outperform face recognition ones.[9]

While it has been suggested that, among the existing biometrics traits, ear biometrics are considered very viable and robust in line with iris and fingerprint biometrics,[35] regarding ear recognition and recognition based on fingerprints, it is clear that fingerprints significantly outperform ear recognition, even when considering novel methods of ear recognition.[19][24][30][33][35][39][41][45] This could perhaps be attributed to fingerprints having more unique feature. However, as noted before, ear characteristics may be as permanent as fingertips and equally, or more difficult, to counterfeit.

It should also be noted that multimodal recognition techniques, making use of ear biometrics in combination with other biometric data almost always result in statistically significant improvements over any technique based on any of the individual biometrics.

There have also been notable advances towards a fully automated recognition of the ear, with different approaches being noted regarding the shape and the size of the acquisition window to be used and the different angles employed in image and sound acquisition. Furthermore, such issues as the elasticity and the deformation of the ear can now be addressed to some extent, through the employment of different types of ear biometrics.

Ear biometrics mostly involve non-invasive acquisition techniques, and while age may significantly affect the ears, their characteristics are adequately permanent. Furthermore, ear characteristics are hard to counterfeit and potential culprits are not yet accustomed to ear recognition being used for identification and verification purposes, especially in any forensic science applications. Moreover, ear biometrics are not volatile and thus are inherently difficult to successfully spoof.

Finally, such issues as the uniqueness of ears and of their characteristics, the relation between characteristics of a person's two ears, new ways to fully address ear occlusion and pose variations, and new means and techniques to achieve correct and adequate automatic ear registration and recognition remain open for future scientific research.[1][9] However, all the above-mentioned facts indicate that the scientific field of ear biometrics is at the moment not only viable,[4] but also really thriving.